\documentclass[letterpaper, 10 pt, conference]{ieeeconf}  

\IEEEoverridecommandlockouts                              

\usepackage{amsmath} 

\usepackage{graphicx}
\usepackage{makecell}
\usepackage{threeparttable} 
\usepackage{subcaption}
\usepackage[table]{xcolor}

\newif\ifproofread
\proofreadfalse

\newcommand{\changemarker}[1]{%
\ifproofread
{\color{red}{#1}}%
\else
#1%
\fi
}

\title{\LARGE \bf
GPS-aided Visual Wheel Odometry
}

\author{Junlin Song$^{1}$, Pedro J. Sanchez-Cuevas$^{2}$, Antoine Richard$^{1}$ and Miguel Olivares-Mendez$^{1}$
\thanks{This research was supported by the European Union’s Horizon 2020 project SESAME (grant agreement No 101017258). 
}
\thanks{$^{1}$ Space Robotics (SpaceR) Research Group, Int. Centre for Security, Reliability and Trust (SnT), University of Luxembourg, Luxembourg.} 
\thanks{$^{2}$Advanced Centre for Aerospace Technologies (CATEC) Seville, Spain.} 
}

\begin{document}

\maketitle
\thispagestyle{empty}
\pagestyle{empty}

\begin{abstract}

This paper introduces a novel GPS-aided visual-wheel odometry (GPS-VWO) for ground robots. The state estimation algorithm tightly fuses visual, wheeled encoder and GPS measurements in the way of Multi-State Constraint Kalman Filter (MSCKF). To avoid accumulating calibration errors over time, the proposed algorithm calculates the extrinsic rotation parameter between the GPS global coordinate frame and the VWO reference frame online as part of the estimation process. The convergence of this extrinsic parameter is guaranteed by the observability analysis and verified by using real-world visual and wheel encoder measurements as well as simulated GPS measurements. \changemarker{Moreover, a novel theoretical finding is presented that the variance of unobservable state could converge to zero for specific Kalman filter system.} We evaluate the proposed system extensively in large-scale urban driving scenarios. 
\changemarker{The results demonstrate that better accuracy than GPS is achieved through the fusion of GPS and VWO.}
The comparison between extrinsic parameter calibration and non-calibration shows significant improvement in localization accuracy thanks to the online calibration.

\end{abstract}


\section{INTRODUCTION}

The localization algorithm of mobile ground robots is an essential component in critical industries like industrial automation, autonomous driving and even planetary exploration. In the last two decades, the research community has been looking into high-precision and robust localization algorithms to enable the exploitation of autonomous vehicles to improve the efficiency and productivity of those industries. Typical sensor setups use camera, inertial measurement unit (IMU) and wheels' position encoder (also referred to as odometer in this paper) to solve this localization problem. It is well known that the fusion of camera and IMU measurements can provide accurate motion tracking. This 6-degrees of freedom (DoF) localization technique is called visual-inertial odometry (VIO). Previous studies have shown that the scale information of VIO system is not observable when the robot undergoes constant acceleration motion \cite{martinelli2011vision}\cite{wu2017vins}. Some approaches additionally fuse wheel encoder measurements overcome this scale issue, as wheel-encoders can provide valid scale information \cite{wu2017vins}\cite{liu2019visual}. 

We are interested in the fusion of camera and wheel encoder called visual-wheel odometry (VWO) to deal with situation where the IMU is unavailable or too noisy to be used, or the system undergoes degenerated motion profile.
Thus the system reliability is increased. Compared with the IMU, the wheel-encoders can directly obtain the velocity, thus avoiding the double integration of the acceleration from the IMU. Moreover, its measurements are not affected by biases, unlike the IMU. The localization result obtained by any previously mentioned odometry is doomed to drift, 
as these sensors only generate relative motion constraints. A straightforward strategy to prevent this drifting is to integrate global position measurements such as GPS. 

To correlate GPS measurements with the position output of the VWO system, it is necessary to calculate or estimate the spatial transformation of the two reference frames, referred to as extrinsic parameters in the rest of the paper. In this study, we propose a tightly-coupled filter-based GPS-VWO system with online spatial calibration to fully exploit the information of those sensors. 
The observability analysis of the extrinsic parameters linked to the spatial calibration shows that it is reasonable to add this calibration as an observable variable to the state vector. The online correction of the spatial calibration improves the localization accuracy and avoids dragging large uncertainties over time. The contributions of this work are summarized as:

\begin{itemize}
    \item To the best of our knowledge, this is the first time that a tightly-coupled filter-based GPS-VWO system is proposed to optimally fuse camera, wheel encoder and GPS measurements. 
    The extrinsic parameter between GPS frame and VWO frame is calibrated online.
    
    \item Nonlinear observability theory is employed to analyze the observability of extrinsic parameter between the GPS global coordinate frame and the VWO reference frame. 
    We validate the observability conclusion through simulation experiments.

    \changemarker{\item Observability analysis of extrinsic parameter reveals a novel theoretical finding that the variance of unobservable state could converge to zero. We provide a detailed proof for a simplified Kalman filter system.}
    
    \item The proposed GPS-VWO system is evaluated on large-scale urban driving datasets. The drift of VWO is significantly reduced by GPS. Meanwhile, the system is resilient to GPS interruption and challenge GPS noise (range from 6m to 20m).
    
    \item Moreover, real experiments show that VIO easily suffers from constant velocity motion, while our VWO not. Compared to the state-of-the-art GPS-VIO \cite{lee2020intermittent}, GPS-VWO achieves up to 66\% accuracy improvements.
     
\end{itemize}



\section{Related Work}

High-precision and drift-free localization is essential for the autonomy of ground robots. In recent years, tightly-coupled VIO approaches fusing measurements from camera and IMU have shown great success \cite{huang2019visual}. Practically, VIO algorithms can be classified into two branches, one is based on nonlinear optimization such as VINS-Mono \cite{qin2018vins} and ORB-SLAM3 \cite{campos2021orb}, and the other is based on Kalman Filter such as the commonly used algorithm named Multi-State Constraint Kalman Filter (MSCKF) \cite{mourikis2007multi, li2013high}. The MSCKF-based VIO algorithm is appealing for platforms with limited computing resources, since it consumes much less computational resources, but yet obtains comparable accuracy with optimization-based competitors \cite{geneva2020openvins}.

However, in some specific motion profiles, VIO algorithms can suffer from unobservability \cite{martinelli2011vision}, that is the impossibility to observe specific elements of the state. A relevant case leading to unobservable scale is constant acceleration motion, which is very common for ground robot \cite{wu2017vins}. A detailed motion degradation analysis of VIO is also presented in \cite{yang2019degenerate}. Therefore, it is necessary to consider additional sensor modality to resolve this issue. Wheel-odometers can provide velocity and scale information. Hence, it makes sense to combine camera, IMU and wheel-encoders measurements in a tightly-coupled fashion \cite{wu2017vins, liu2019visual, lee2020visual}. A recent work \cite{zhang2021pose} adapts the above strategy through the optional usage of IMU measurements. In \cite{zhang2021pose}, the motion manifold of ground robots is modeled in detail to improve the localization accuracy.


Compared with the three sensors mentioned above, GPS is able to provide drift-free absolute position information directly. In order to achieve high-quality global localization, it is natural to integrate GPS measurements with those sensor suits or their subsets in large-scale environments. As eluded to earlier, the focus of this paper is the tightly-coupled fusion of GPS measurements in a VWO system, which is not studied in the literature. 
First, compared to loosely-coupled algorithms \cite{qin2019general}, tightly-coupled algorithms provides higher localization accuracy, because tightly-coupled algorithms fully exploited the inner correlation of all measurements \cite{cioffi2020tightly}. Second, in order to realize multi-sensor fusion, we need to calculate extrinsic parameter to model GPS measurements \cite{lee2020intermittent}. 
Third, it is critical to prove that the extrinsic parameter is observable before including this parameter to state vector \cite{lee2020intermittent}. If this parameter is observable, performing online estimation can reduce the initialization error, which helps to improve the accuracy of the estimator output. If not, it will deteriorate the estimation. Lie derivative is a readily-available tool for the observability analysis of nonlinear system \cite{hermann1977nonlinear}. This tool was previously used to analyze the observability of extrinsic parameter between camera and IMU in VIO systems \cite{mirzaei2008kalman, kelly2011visual}. Here, we employ Lie derivative to analyze the observability of extrinsic parameter between the GPS reference frame and the VWO reference frame, and drawn non-trivial conclusion, which is different from \cite{lee2020intermittent}.

\section{Problem Formulation}

\subsection{Notation}

\begin{figure}[htbp]
  \centering
    \begin{subfigure}[t]{0.23\textwidth}
        \centering
        \includegraphics[width=\textwidth, height=0.8\textwidth]{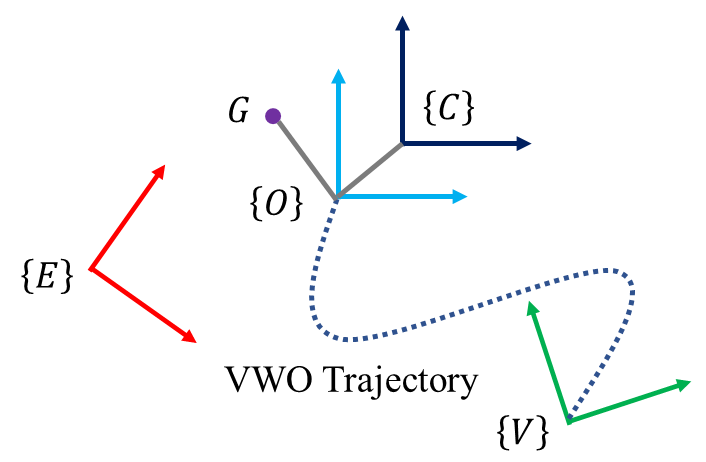}
        \caption{}
        \label{Frames}
    \end{subfigure}
    \hfill
    \begin{subfigure}[t]{0.23\textwidth}
        \centering
        \includegraphics[width=\textwidth, height=0.8\textwidth]{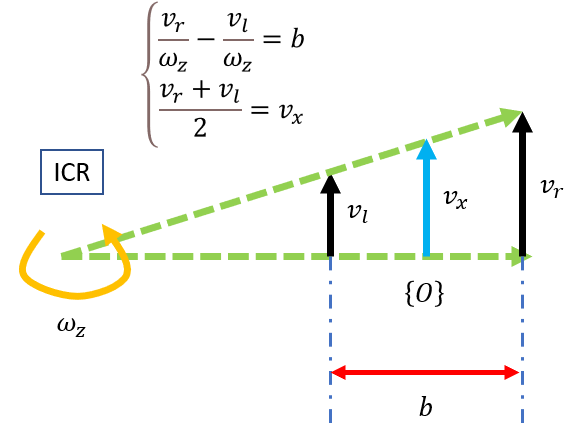}
        \caption{}
        \label{wheel}
    \end{subfigure}
  \caption{(a) Coordinate frames. (b) The relationship between the velocity of two differential (left and right) wheels.}
  \label{Notation}
\end{figure}

The coordinate frames used in this paper are illustrated in the Fig. \ref{Frames}. $\{ E\} $  is the East-North-Up (ENU) coordinate frame. The GPS measurements are described in this frame. Usually, the first received GPS measurement is selected as the origin of  $\{ E\} $. $\{ V\} $  denotes the reference frame of the VWO system. $\{ O\} $  and  $\{ C\} $ denote the wheel-odometer frame and camera frame respectively. $G$  represents the position of the receiving antenna of GPS. The studied vehicle follows the differential drive kinematics. Fig. \ref{wheel} shows the relationship between the velocity of two differential (left and right) wheels. $v_l$  and $v_r$  are the velocity of left and right wheel respectively. The ICR represents the instantaneous center of rotation. The origin of $\{ O\} $ is  located at the center of the robot's wheels. $b$ is the distance between the wheels.


We use ${}^V\left(  \bullet  \right)$  to represent a physical quantity in the frame $\{ V\} $. The position of a point $O$ in the frame $\{ V\}$ is expressed as ${}^V{p_O}$. The velocity of a point $O$ in the frame $\{ V\}$ is expressed as ${}^V{v_O}$. A Unit quaternion is employed to represent the rotation of the rigid body \cite{trawny2005indirect}. ${}_V^Oq$ represents the orientation of the frame $\{ O\}$ with respect to the frame $\{ V\} $, and its corresponding rotation matrix is ${}_V^OR$. $\left\{ {\begin{array}{*{20}{c}}
{{}_O^CR}&{{}^C{p_O}}
\end{array}} \right\}$ is the extrinsic parameter set between camera and wheel encoder. ${}^O{p_G}$ is the extrinsic parameter from the GPS sensor to frame $\{ O\} $. These extrinsic parameters can be obtained from CAD model or calibrated before system running. This paper assumes that they are known. ${\left[  \bullet  \right]_ \times }$ is denoted as the skew symmetric matrix corresponding to a three-dimensional vector. The transpose of a matrix is ${\left[  \bullet  \right]^T}$. $e_i$ is a $3 \times 1$ vector, with the $i$th element to be 1 and other elements to be 0.


\subsection{MSCKF based VWO} \label{VWO}
In this section, we first briefly review the pipeline of standard MSCKF \cite{mourikis2007multi}. The classical state estimation algorithm based on MSCKF usually defines a state vector containing the current state and multiple historical state clones:

\begin{equation}
    \begin{array}{l}
    {x_O} = {\left( {\begin{array}{*{20}{c}}
    {{}_V^O{q^T}}&{{}^Vp_O^T}&{{}^Vp_f^T}
    \end{array}} \right)^T}\\
    {x_{{c_i}}} = {\left( {\begin{array}{*{20}{c}}
    {{}_V^{{O_i}}{q^T}}&{{}^Vp_{{O_i}}^T}
    \end{array}} \right)^T}\\
    x = {\left( {\begin{array}{*{20}{c}}
    {x_O^T}&{x_{{c_1}}^T}& \cdots &{x_{{c_N}}^T}
    \end{array}} \right)^T}
    \end{array}
\end{equation}

Where $x_O$ includes the position and orientation of the wheel-odometer frame $\{ O\} $ at the current time and the visual feature position, ${}^V{p_f}$, in the reference frame of the VWO. To be concise, only one visual feature point is described here, namely $f$. This simplified case has no impact
on the subsequent theoretical analysis. ${x_{{c_i}}}$  is obtained by cloning the first two physical quantities of $x_O$  at different times. Then the state vector of the VWO system can be obtained by augmenting $x_O$ with $N$ historical clones of ${x_{{c_i}}}$ . $N$ is the sliding window size, a fixed parameter.

\subsubsection{Wheel Odometer Propagation}\hfill\\
We propagate the state and covariance based on the kinematic model of a differential wheels vehicle taken from \cite{siegwart2011introduction, censi2013simultaneous}. The wheeled encoder equipped by the ground-robot provides local linear and angular velocity measurement (see Fig. \ref{wheel}):

\begin{equation}
    \begin{array}{l}
    {v_l} = \frac{{\Delta {m_l} + {n_l}}}{{{M_l}\Delta t}}\pi {d_l}\\
    {v_r} = \frac{{\Delta {m_r} + {n_r}}}{{{M_r}\Delta t}}\pi {d_r}\\
    {v_x} = \frac{{{v_l} + {v_r}}}{2}\\
    {\omega _z} = \frac{{{v_r} - {v_l}}}{b}
    \end{array}
\end{equation}

Where $\Delta {m_l}$ is the reading difference between two adjacent measurements of left wheel's encoder. Time difference between these two adjacent timestamps is denoted as $\Delta t$.
The meaning of $\Delta {m_r}$  is the same as $\Delta {m_l}$.
${n_l}$  and ${n_r}$ are zero mean Gaussian noise corresponding to $\Delta {m_l}$ and $\Delta {m_r}$ respectively. ${M_l}$ and ${M_r}$ represent the resolution of left and right wheels' encoders respectively. $d_l$ and $d_r$ represent the left and right wheels' diameters respectively. The above measurements can be used to compose the linear and angular velocity in frame $\{ O\} $ :

\begin{equation}
    {}^O{\omega _O} = \left[ {\begin{array}{*{20}{c}}
    {{n_{{\omega _x}}}}\\
    {{n_{{\omega _y}}}}\\
    {{\omega _z}}
    \end{array}} \right]{\rm{  }}{}^O{v_O} = \left[ {\begin{array}{*{20}{c}}
    {{v_x}}\\
    {{n_{{v_y}}}}\\
    {{n_{{v_z}}}}
    \end{array}} \right]
\end{equation}

Where ${n_{\left[ \bullet \right]}}$  represents the zero mean Gaussian noise of $\left[  \bullet \right]$. The continuous-time kinematic model of wheel-odometer is given by:

\begin{equation}
    \begin{array}{l}
    {}_V^O\dot q = \frac{1}{2}\Omega \left( {{}^O{\omega _O}} \right){}_V^Oq\\
    {}^V{{\dot p}_O} = {}_V^O{R^T}{}^O{v_O}\\
    \end{array}
\end{equation}

Meanwhile,

\begin{equation}
    \Omega \left( \omega  \right) = \left[ {\begin{array}{*{20}{c}}
    { - {{\left[ \omega  \right]}_ \times }}&\omega \\
    { - {\omega ^T}}&0
    \end{array}} \right]
\end{equation}

\subsubsection{Visual Measurement Update}\hfill\\
We refer to Open-VINS \cite{geneva2020openvins} for the technical details of the visual measurement update. When a new image comes, the latest state is cloned and the state vector is augmented by this clone state. Then, the carefully selected feature points are utilized to update the poses over the sliding window and the position of the feature points, through camera measurement model and triangulation.

\subsection{GPS Measurement Update}
The GPS position measurement is expressed in the frame $\{ E\} $ . The measurement equation is modelled as \footnote{Measurement equation (\ref{eq:gps}) is used here just for the convenience of the observability analysis. In the implementation, we adopt the interpolation measurement equation as \cite{lee2020intermittent}.}:

\begin{equation}
\begin{aligned}
    {}^E{p_G} &= {}^E{p_V} + {}_V^ER{}^V{p_G}\\
    &= {}^E{p_V} + {}_V^ER( {{}^V{p_O} + {}_V^O{R^T}{}^O{p_G}})
\end{aligned}
\label{eq:gps}
\end{equation}

Where ${}^O{p_G}$ is the extrinsic parameter from the GPS sensor to frame $\{ O\} $. ${}^V{p_O}$  and ${}_V^OR$ are described in section \ref{VWO}. $\left\{ {\begin{array}{*{20}{c}}
{{}_V^ER}&{{}^E{p_V}}
\end{array}} \right\}$  is the extrinsic parameter set between the frame $\{ V\} $  and the frame $\{ E\} $ .

The parameterization of ${}_V^ER$  will be discussed in detail. This paper mainly analyzes two parameterization methods. One is a 3DoF parameterization, while the other is 1DoF. One attitude angle from yaw, pitch, or roll can be selected as \textbf{“active DoF”}, while others are denoted as \textbf{“fixed DoF”}. For example, if yaw is used as "active DoF", ${}_V^ER$ can be expressed in the following form:

\begin{equation}
\begin{aligned}
    {}_V^ER &= {R_z}\left( \alpha  \right){R_y}\left( \beta  \right){R_x}\left( \gamma  \right)\\
    &= \left[ {\begin{array}{*{20}{c}}
    {\cos \alpha }&{ - \sin \alpha }&0\\
    {\sin \alpha }&{\cos \alpha }&0\\
    0&0&1
    \end{array}} \right]{R_y}\left( \beta  \right){R_x}\left( \gamma  \right)
\end{aligned}
\end{equation}

Where $\alpha$, $\beta$ and $\gamma$ are yaw , pitch and roll respectively. However, $\beta$ and $\gamma$ are fixed, marked as "fixed DoF". $\alpha$ is the only variable can be changed during the update, marked as "active DoF", thus 1DoF.


To analyse the observability of the extrinsic parameter between the frame $\{ V\} $ and the frame $\{ E\} $, this parameter is included to the state vector.

The new system state vector becomes:

\begin{equation}
    x = {\left( {\begin{array}{*{20}{c}}
    {x_O^T}&{x_{{c_1}}^T}& \cdots &{x_{{c_N}}^T}&{{}_V^E{q^T}}&{{}^Ep_V^T}
    \end{array}} \right)^T}
\end{equation}

The subset of state variables related to the GPS measurement equation is noted as ${x_s}$ :

\begin{equation}
    {x_s} = {\left[ {\begin{array}{*{20}{c}}
    {{}_V^O{q^T}}&{{}^Vp_O^T}&{{}_V^E{q^T}}&{{}^Ep_V^T}
    \end{array}} \right]^T}
\end{equation}

The corresponding error state is expressed as:

\begin{equation}
    {\tilde x_s} = {\left[ {\begin{array}{*{20}{c}}
    {{}_V^O{{\tilde \theta }^T}}&{{}^V\tilde p_O^T}&{{}_V^E{{\tilde \theta }^T}}&{{}^E\tilde p_V^T}
    \end{array}} \right]^T}
\end{equation}

The measurement Jacobian of GPS is calculated as:

\begin{equation}
\begin{aligned}
    H &= \frac{{\partial  {}^E{\tilde p_G}}}{{\partial {{\tilde x}_s}}}\\
    &= \left[ {\begin{array}{*{20}{c}}
    { - {}_V^E R{}_V^O{{ R}^T}{{\left[ {{}^O{p_G}} \right]}_ \times }}&{{}_V^E R}&{{H_\theta }}&{{I_3}}
    \end{array}} \right]
\end{aligned}
\end{equation}

If ${}_V^E\theta $ is a 3DoF parameterization:

\begin{equation} \label{eq: 3dof_jacob}
    {H_\theta } = {\left[ {{}_V^E R{}^V{{ p}_G}} \right]_ \times }
\end{equation}

If ${}_V^E\theta $ is a 1DoF parameterization:

\begin{equation}
    {H_\theta } = \left[ {\begin{array}{*{20}{c}}
    { - \sin \alpha }&{ - \cos \alpha }&0\\
    {\cos \alpha }&{ - \sin \alpha }&0\\
    0&0&0
    \end{array}} \right]{R_y}\left( \beta  \right){R_x}\left( \gamma  \right){}^V{ p_G}
\end{equation}

We omit the initialization of GPS-VWO system and the proper handling of time offset among multi-sensors, because these are not the main focus of this work. Interested readers can refer to \cite{lee2020intermittent}.

\section{Observability Analysis} \label{Observability Analysis}
The observability plays an important role in state estimation.  In this section, we use Lie derivatives to perform the observability analysis of a nonlinear system \cite{hermann1977nonlinear}. To the best of our knowledge, this is the first time that a paper studies the observability of the extrinsic parameter of a GPS-VWO system. Subsequently, We will prove that the translation part of the extrinsic parameter is unobservable and the rotation part of the extrinsic parameter is unobservable when it is parameterized with 3DoF, yet observable when it is parameterized with 1DoF.

To simplify the formulation, the cloned states in the state vector are removed. The system state vector now becomes:

\begin{equation}
    x = {\left[ {\begin{array}{*{20}{c}}
    {{}_V^O{q^T}}&{{}^Vp_O^T}&{{}^Vp_f^T}&{{}_V^E{\theta ^T}}&{{}^Ep_V^T}
    \end{array}} \right]^T}
\end{equation}

The kinematic equation is rewritten into a suitable form to calculate the Lie derivatives.

\begin{equation}
    \dot x = \left[ {\begin{array}{*{20}{c}}
    {{}_V^O\dot q}\\
    {{}^V{{\dot p}_O}}\\
    {{}^V{{\dot p}_f}}\\
    {{}_V^E\dot \theta }\\
    {{}^E{{\dot p}_V}}
    \end{array}} \right] = \underbrace {\left[ {\begin{array}{*{20}{c}}
    {\frac{1}{2}\Xi \left( {{}_V^Oq} \right)}\\
    {{0_3}}\\
    {{0_3}}\\
    {{0_{D \times 3}}}\\
    {{0_3}}
    \end{array}} \right]}_{{f_1}}{}^O{\omega _O} + \underbrace {\left[ {\begin{array}{*{20}{c}}
    {{0_3}}\\
    {{}_V^O{R^T}}\\
    {{0_3}}\\
    {{0_{D \times 3}}}\\
    {{0_3}}
    \end{array}} \right]}_{{f_2}}{}^O{v_O}
\end{equation}

The value of $D$ is either 3 or 1 depending on the number of DoF. In addition, here we use the following property of time derivative of quaternion:

\begin{equation}
    \dot q = \frac{1}{2}\Omega \left( \omega  \right)q = \frac{1}{2}\Xi \left( q \right)\omega
\end{equation}

The definition of $\Xi \left( q \right)$  can be found in \cite{trawny2005indirect}.
Next, we list the usable measurement equations. The camera measurement equation is:

\begin{equation}
    {h_1}\left( x \right) = {}^C{p_f} = {}_O^CR{}_V^ORp + {}^C{p_O}
\end{equation}

Where $p = {}^V{p_f} - {}^V{p_O}$ .

The norm constraint of the unit quaternion is also considered as its own measurement equation:

\begin{equation}
    {h_2}\left( x \right) = {}_V^O{q^T}{}_V^Oq - 1 = 0
\end{equation}

The measurement equation of GPS is:

\begin{equation}
    {h_3}\left( x \right) = {}^E{p_G} = {}^E{p_V} + {}_V^ER{}^V{p_O}
\end{equation}

If the original equation (\ref{eq:gps}) is adopted, the same analysis process will generate consistent conclusions. Therefore, assumption ${}^O{p_G} = {0_{3 \times 1}}$  is made here to simplify the description.

\subsubsection{Zeroth-Order Lie Derivatives}\hfill\\
The zeroth-order Lie derivative of a function is itself.

\begin{equation}
    \begin{array}{l}
    {\pounds^0}{h_1} = {}^C{p_O} + {}_O^CR{}_V^ORp\\
    {\pounds^0}{h_2} = {}_V^O{q^T}{}_V^Oq - 1\\
    {\pounds^0}{h_3} = {}^E{p_V} + {}_V^ER{}^V{p_O}
    \end{array}
\end{equation}

The gradients of the zeroth-order Lie derivatives with respect to $x$ are:

\begin{equation}
    \begin{array}{l}
    \nabla {\pounds^0}{h_1} = \left[ {\begin{array}{*{20}{c}}
    {\Psi }&{ - {}_O^CR{}_V^OR}&{{}_O^CR{}_V^OR}&{{0_{3 \times D}}}&{{0_3}}
    \end{array}} \right]\\
    \nabla {\pounds^0}{h_2} = \left[ {\begin{array}{*{20}{c}}
    {2{}_V^O{q^T}}&{{0_{1 \times 3}}}&{{0_{1 \times 3}}}&{{0_{1 \times D}}}&{{0_{1 \times 3}}}
    \end{array}} \right]\\
    \nabla {\pounds^0}{h_3} = \left[ {\begin{array}{*{20}{c}}
    {{0_{3 \times 4}}}&{{}_V^ER}&{{0_3}}&{{H_\theta }}&{{I_3}}
    \end{array}} \right]
    \end{array}
\end{equation}

Meanwhile,

\begin{equation}
    \Psi
    = {}_O^CR\frac{{\partial R\left( {}_V^Oq \right)p}}{{\partial {}_V^Oq}}
\end{equation}

\subsubsection{First-Order Lie Derivatives}\hfill\\
The first-order Lie derivative of $h_1$ with respect to $f_1$ is computed as:

\begin{equation}
    \pounds_{{f_1}}^1{h_1} = \nabla {\pounds^0}{h_1} \bullet {f_1} = \frac{1}{2}\Psi 
    \Xi \left( {{}_V^Oq} \right){e_3}
\end{equation}

The gradient of $\pounds_{{f_1}}^1{h_1}$ with respect to $x$  is: 

\begin{equation}
    \nabla \pounds_{{f_1}}^1{h_1} = \left[ {\begin{array}{*{20}{c}}
    {\Gamma }&{ - \Upsilon }&{\Upsilon}&{{0_{3 \times D}}}&{{0_3}}
    \end{array}} \right]
\end{equation}

Where $\Gamma $ and $\Upsilon $ do not need to be explicitly computed because they do not affect the observability analysis.

The first-order Lie derivative of $h_3$  with respect to $f_2$  is computed as:

\begin{equation}
    \pounds_{{f_2}}^1{h_3} = \nabla {\pounds^0}{h_3} \bullet {f_2} = {}_V^ER{}_V^O{R^T}{e_1}
\end{equation}

The gradient of $\pounds_{{f_2}}^1{h_3}$  with respect to $x$  is: 

\begin{equation}
    \nabla \pounds_{{f_2}}^1{h_3} = \left[ {\begin{array}{*{20}{c}}
    X&{{0_3}}&{{0_3}}&Y&{{0_3}}
    \end{array}} \right]
\end{equation}

Where $X$ is not explicitly computed as $\Gamma $ and $\Upsilon $. $Y$  is:

\begin{equation}
    Y
    = \frac{{\partial {}_V^ER{}_V^O{R^T}{e_1}}}{{\partial {}_V^E\theta }}
\end{equation}

If ${}_V^E\theta$ is a 3DoF parameterization:

\begin{equation}
    Y
    = {\left[ {{}_V^ER{}_V^O{R^T}{e_1}} \right]_ \times }
\end{equation}

If ${}_V^E\theta$ is a 1DoF parameterization:

\begin{equation}
    Y
    = \left[ {\begin{array}{*{20}{c}}
    { - \sin \alpha }&{ - \cos \alpha }&0\\
    {\cos \alpha }&{ - \sin \alpha }&0\\
    0&0&0
    \end{array}} \right]{R_y}\left( \beta  \right){R_x}\left( \gamma  \right){}_V^O{R^T}{e_1}
\end{equation}

\subsubsection{observability analysis}\hfill\\
By stacking the gradients of previously calculated Lie derivatives together, the following observability matrix is constructed:
\begin{equation}
    \scalebox{0.9}{$
    \begin{array}{c}
    {\cal O} = \left[ {\begin{array}{*{20}{c}}
    {\nabla {\pounds^0}{h_1}}&
    {\nabla {\pounds^0}{h_2}}&
    {\nabla {\pounds^0}{h_3}}&
    {\nabla \pounds_{{f_1}}^1{h_1}}&
    {\nabla \pounds_{{f_2}}^1{h_3}}
    \end{array}} \right]^T
    \\ = 
    \left[ {\begin{array}{*{20}{c}}
    {\Psi }&{ - {}_O^CR{}_V^OR}&{{}_O^CR{}_V^OR}&{{0_{3 \times D}}}&{{0_3}}\\
    {2{}_V^O{q^T}}&{{0_{1 \times 3}}}&{{0_{1 \times 3}}}&{{0_{1 \times D}}}&{{0_{1 \times 3}}}\\
    {{0_{3 \times 4}}}&{{}_V^ER}&{{0_3}}&{{H_\theta }}&{{I_3}}\\
    {\Gamma }&{ - \Upsilon }&{\Upsilon }&{{0_{3 \times D}}}&{{0_3}}\\
    X&{{0_3}}&{{0_3}}&Y&{{0_3}}
    \end{array}} \right]
    \end{array}
    $}
\end{equation}

Adding the third column to the second column, ${\cal O}$  becomes:
\begin{equation}
    \scalebox{0.9}{$
    {\cal O} = \left[ {\begin{array}{*{20}{c}}
    {\Psi }&{{0_3}}&{{}_O^CR{}_V^OR}&{{0_{3 \times D}}}&{{0_3}}\\
    {2{}_V^O{q^T}}&{{0_{1 \times 3}}}&{{0_{1 \times 3}}}&{{0_{1 \times D}}}&{{0_{1 \times 3}}}\\
    {{0_{3 \times 4}}}&{{}_V^ER}&{{0_3}}&{{H_\theta }}&{{I_3}}\\
    {\Gamma }&{{0_3}}&{\Upsilon }&{{0_{3 \times D}}}&{{0_3}}\\
    X&{{0_3}}&{{0_3}}&Y&{{0_3}}
    \end{array}} \right]
    $}
\end{equation}

\vspace*{1pt}
 ${}_V^ER$ in the second column can be used to eliminate ${H_\theta }$  in the fourth column and  ${I_3}$ in the fifth column. Thus, ${\cal O}$ reduces to: 
 \begin{equation}
    \scalebox{0.9}{$
    {\cal O} = \left[ {\begin{array}{*{20}{c}}
    {\Psi }&{{0_3}}&{{}_O^CR{}_V^OR}&{{0_{3 \times D}}}&{{0_3}}\\
    {2{}_V^O{q^T}}&{{0_{1 \times 3}}}&{{0_{1 \times 3}}}&{{0_{1 \times D}}}&{{0_{1 \times 3}}}\\
    {{0_{3 \times 4}}}&{{}_V^ER}&{{0_3}}&{{0_{3 \times D}}}&{{0_3}}\\
    {\Gamma }&{{0_3}}&{\Upsilon }&{{0_{3 \times D}}}&{{0_3}}\\
    X&{{0_3}}&{{0_3}}&Y&{{0_3}}
    \end{array}} \right]
    $}
\end{equation}

The fifth column corresponds to the translation part of the extrinsic parameter. This column is not full rank, so the translation part is unobservable. Finally, the rotation part of the extrinsic parameter is analyzed. Let us focus on $Y$ in the fourth column. If ${}_V^E\theta$ is a 3DoF parameterization, $Y$  is a skew symmetric matrix, which means the fourth column is not full rank. Therefore, the rotation part is unobservable. If ${}_V^E\theta$ is a 1DoF parameterization, $\left\| Y \right\| \ne 0$ is hold in general case. The fourth column is full rank and cannot eliminated by other columns. So the rotation part is observable. It is worth noting that although yaw is used as "active DoF" here, the above observability conclusion is also applicable to the pitch or roll case.

\section{Experiments}
To compare the performance of the different algorithms, we adapt our main implementation of GPS-VWO system into three variants. When GPS is not used, the algorithm degenerates to \textbf{“VWO”}. When fusing GPS, we further distinguish the algorithm according to whether to calibrate the extrinsic parameter online.
If the strategy in \cite{lee2020intermittent} is adopted, the algorithm is referred as \textbf{“GPS-VWO-fixed”}, which means the extrinsic parameter is \changemarker{estimated during the initialization stage, then} fixed and marginalized after initialization. If the rotation part of the extrinsic parameter is 1DoF parameterization and calibrated online \changemarker{after initialization},
the algorithm is referred as \textbf{“GPS-VWO”}. 

We verify the observability conclusion derived for GPS-VWO system with simulated Gaussian GPS measurements. Then, all three implementations are evaluated on the large-scale KAIST dataset which features challenging urban driving scenes \cite{jeong2019complex}. The KAIST dataset is collected from a rear-wheel drive passenger car. This dataset includes the following sensors: GPS, camera, and wheel encoders distributed on the left and right rear wheel. The frequencies of GPS, camera, and wheel encoder are 5Hz, 10Hz, and 100Hz respectively. We calculate the absolute trajectory error (ATE) between GPS trajectory and groundtruth trajectory for all datasets. To better present the performance of different algorithms under relatively large GPS noise, datasets with ATE larger than 6m are selected for real-world experiments.

\changemarker{For all experiments, the wheel-odometer noise parameters are set as: ${\sigma _{{n_l}}} = {\sigma _{{n_r}}} = 0.01,{\rm{ }}{\sigma _{{n_{{\omega _x}}}}} = {\sigma _{{n_{{\omega _y}}}}} = 0.01,{\rm{ }}{\sigma _{{n_{{v_y}}}}} = 0.1,{\sigma _{{n_{{v_z}}}}} = 0.01$. The noise of $v_z$ is smaller than that of $v_y$ as the wheel movement in the $z$ direction is constrained, tightly attached to the road surface.}

\subsection{Validation of the Observability Analysis} \label{Validation of the Observability Analysis}

\begin{figure*}[htbp]
  \vspace{16pt}
  \centering
    \begin{subfigure}[t]{0.24\textwidth}
        \centering
        \includegraphics[width=\textwidth, height=1.2\textwidth]{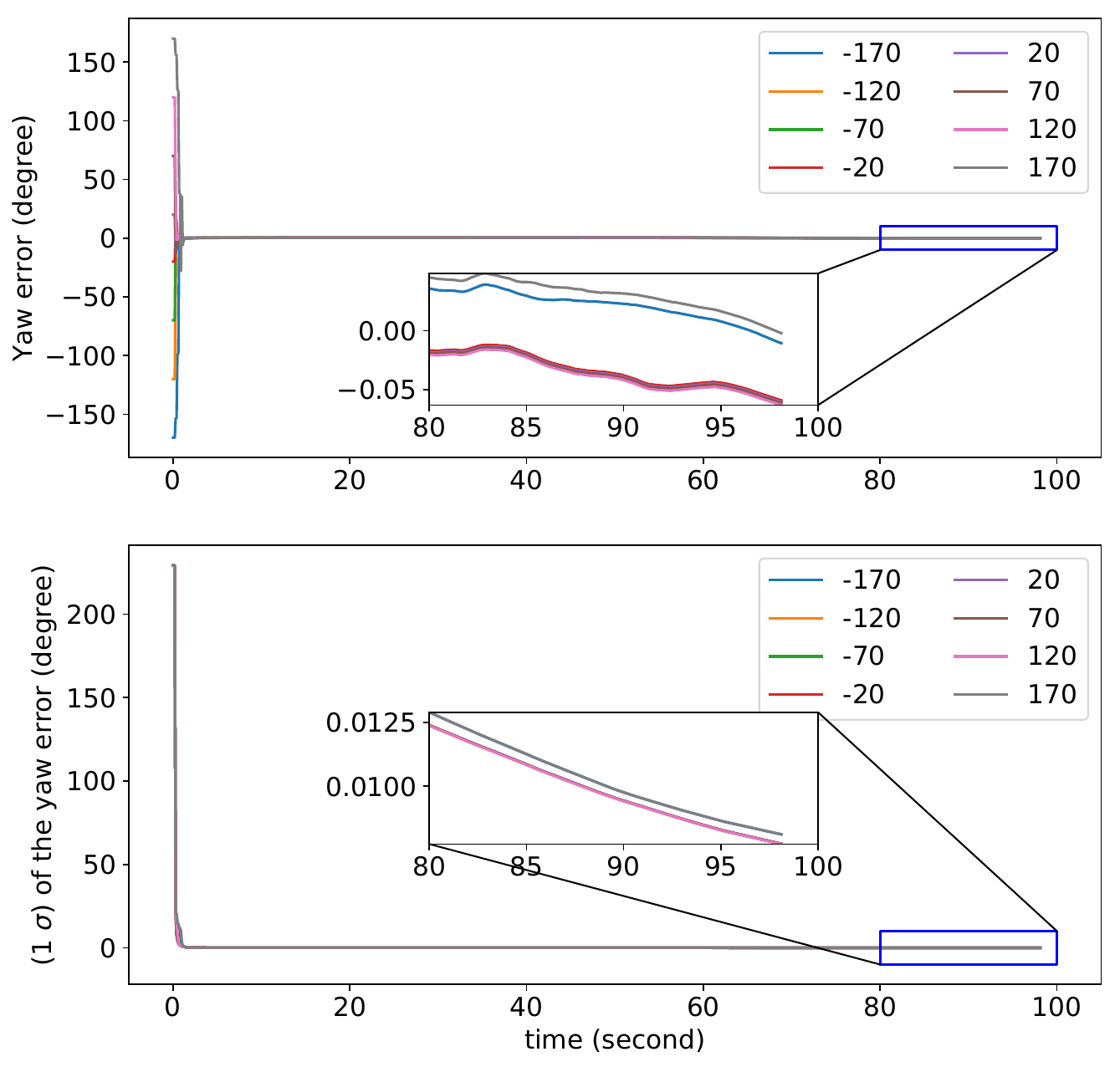}
        \caption{Top: Yaw convergence over time with respect to different initial guesses. Bottom: One standard deviation (1 $\sigma $) of yaw.}
        \label{Yaw merge}
    \end{subfigure}
    \hfill
    \begin{subfigure}[t]{0.24\textwidth}
        \centering
        \includegraphics[width=\textwidth, height=1.2\textwidth]{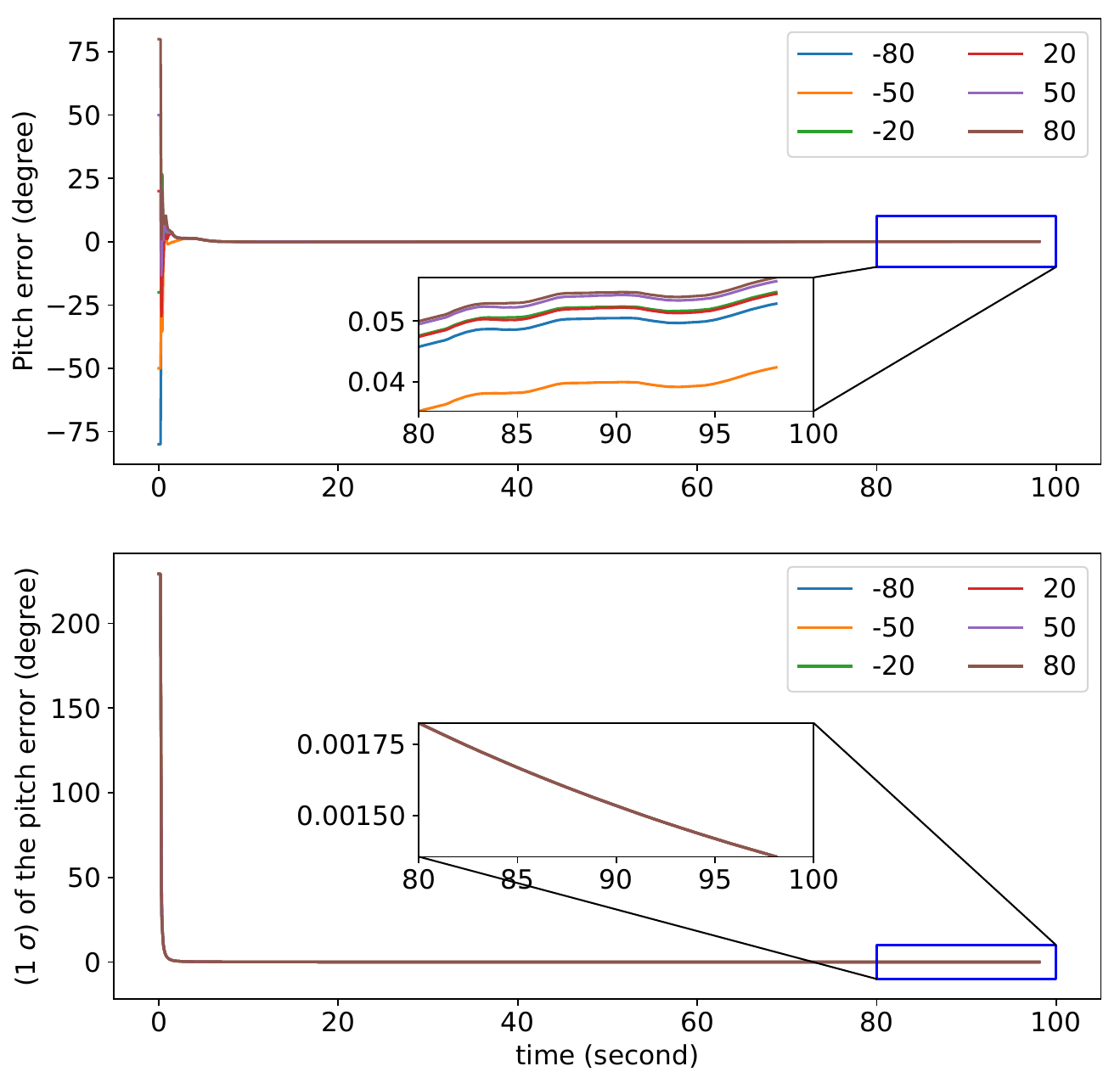}
        \caption{Top: Pitch convergence over time with respect to different initial guesses. Bottom: One standard deviation (1 $\sigma $) of pitch.}
        \label{Pitch merge}
    \end{subfigure}
    \hfill
    \begin{subfigure}[t]{0.24\textwidth}
        \centering
        \includegraphics[width=\textwidth, height=1.2\textwidth]{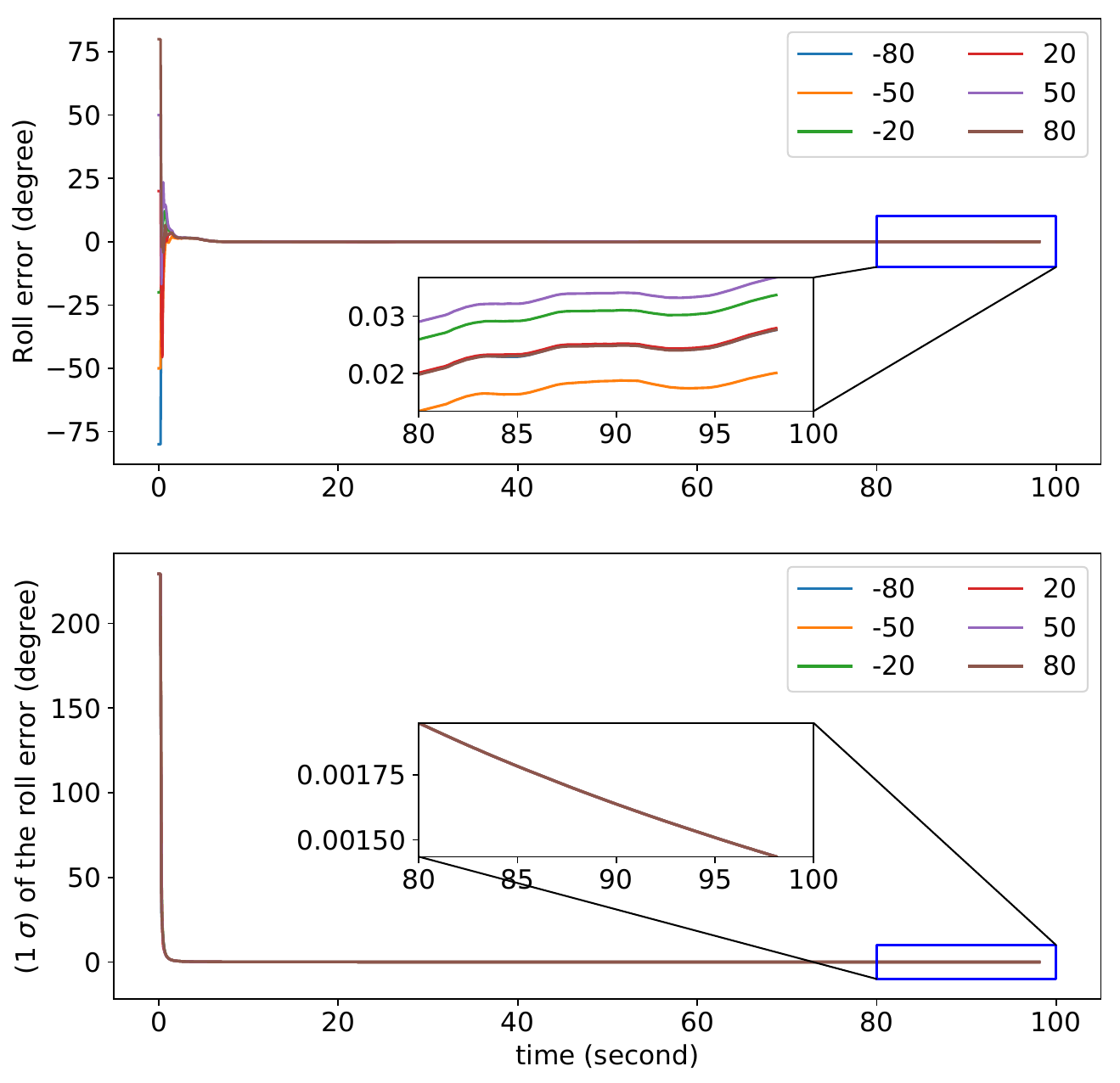}
        \caption{Top: Roll convergence over time with respect to different initial guesses. Bottom: One standard deviation (1 $\sigma $) of roll.}
        \label{Roll merge}
    \end{subfigure}
    \hfill
    \begin{subfigure}[t]{0.24\textwidth}
        \centering
        \includegraphics[width=\textwidth, height=1.2\textwidth]{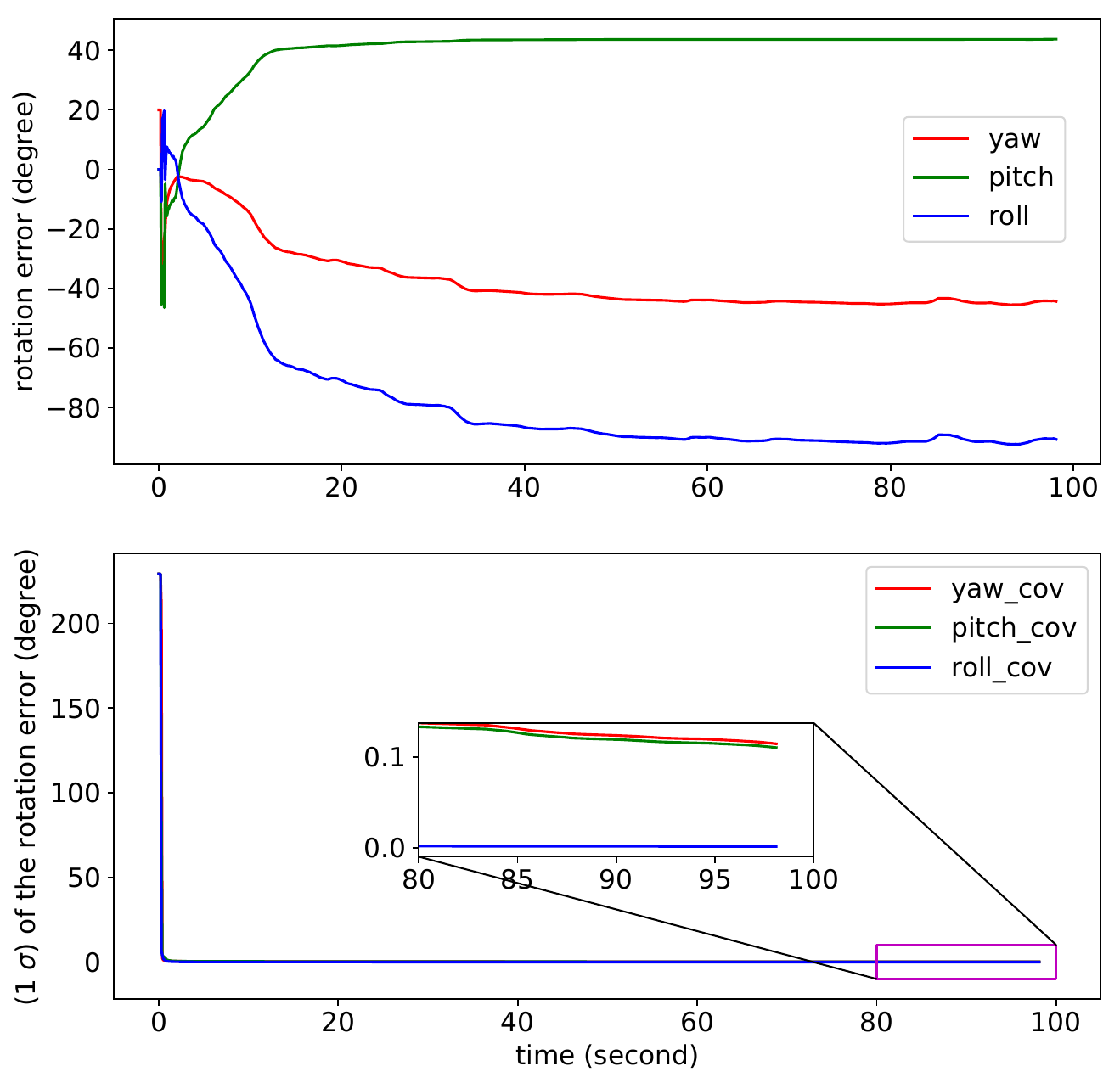}
        \caption{Top: rotation error over time. Bottom: One standard deviation (1 $\sigma $) of rotation error.}
        \label{ypr merge}
    \end{subfigure}
  \caption{Calibration results of different parameterization for rotation extrinsic parameter}
  \label{observability analysis}
\end{figure*}

The experiments in this section are here to validate the observability analysis. 
The KAIST Urban25 dataset is used here. We add zero mean Gaussian noise to the groundtruth position to simulate GPS position measurements. 
Considering GPS noise in vertical is much larger than that in horizontal, the additive Gaussian noise is defined as:

\begin{equation}
    n_{gps} \sim {\cal N}\left( {{0_{3 \times 1}},{diag(1, 1, 4)}} \right)
\end{equation}

The translation part of the extrinsic parameter between the frame $\{ V\} $  and the frame $\{ E\} $  is acquired by querying the groundtruth position values at the initialization time. The rotation part of the extrinsic parameter, ${}_V^ER$ can be obtained in the same way. We add some perturbations to the groundtruth ${}_V^ER$ according to different parameterization and set them as the initial estimate of ${}_V^ER$. 



First, we test if the 1DoF parameterization of ${}_V^ER$ is observable. 
\changemarker{The active DoF is yaw in Fig. \ref{Yaw merge}, pitch in Fig. \ref{Pitch merge} and roll in Fig. \ref{Roll merge}.}
Fig. \ref{Yaw merge} shows the convergence of yaw error and the corresponding one standard deviation (1 $\sigma $) calculated as the square root of the corresponding variance. The range of the initial yaw error is $\left[ { - {{170.0}^ \circ },{{170.0}^ \circ }} \right]$. The estimation of the yaw error consistently converges to near zero with small uncertainty, and the convergence process is robust to the relatively large initial error. Fig. \ref{Pitch merge} shows the convergence of pitch error and its (1 $\sigma $) over time. The range of the initial pitch error is $\left[ { - {{80.0}^ \circ },{{80.0}^ \circ }} \right]$. Similar Results of roll parameterization are shown in Fig. \ref{Roll merge}. From Fig. \ref{Yaw merge} to Fig. \ref{Roll merge}, We can find that when ${}_V^ER$ is a 1DoF parameterization, the rotational extrinsic parameter is able to perfectly approach to groundtruth value in a short period. 

Then, we show that the 3DoF parameterization of ${}_V^ER$ is unobservable. The perturbation angle is set as $\left[ { {{20.0}^ \circ },{{0.0}^ \circ },{{0.0}^ \circ }} \right]$. Fig. \ref{ypr merge} shows the rotation error over time. These three components can not converge to zero. The results support the observability conclusion in section \ref{Observability Analysis}. Regarding to the convergence of the (1 $\sigma $) for each component (see the bottom of Fig. \ref{ypr merge}), we provide an intuitive explanation in APPENDIX.

\begin{figure}[htbp]
  \centering
  \includegraphics[width=0.42\textwidth]{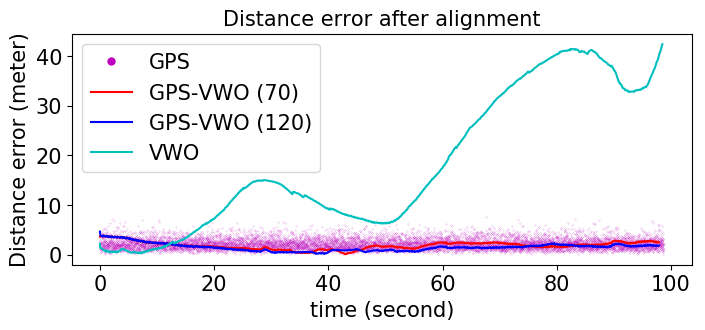}
  \caption{Comparison of distance error with different methods. The initial yaw errors are ${{70.0}^ \circ }$ and ${{120.0}^ \circ }$.}
  \label{Distance error of sim}
\end{figure}

In order to demonstrate the advantage of GPS-VWO in a intuitive fashion, we draw the distance error over time in Fig. \ref{Distance error of sim}. Although the initial yaw error is selected as a relatively large value, the introduction of the GPS significantly reduces the localization error of the VWO. By fusing GPS, the distance error of GPS-VWO is even less than that of the GPS most of the time. 



\subsection{Localization Results with real GPS}
In this section, real GPS measurements provided by the KAIST dataset \cite{jeong2019complex} are used.
We assume that in a ground vehicle, the most important variable necessary to couple the VWO and GPS reference frames is the yaw angle because of the feature that pitch and roll are usually close to zero and prone to be buried by relatively large GPS noise in practice. Moreover, it is effortless to obtain pitch and roll at initialization stage once IMU is utilized, although we do not consider IMU currently. 

\begin{figure}[htbp]
  \centering
  \includegraphics[width=0.48\textwidth, height=0.40\textwidth]{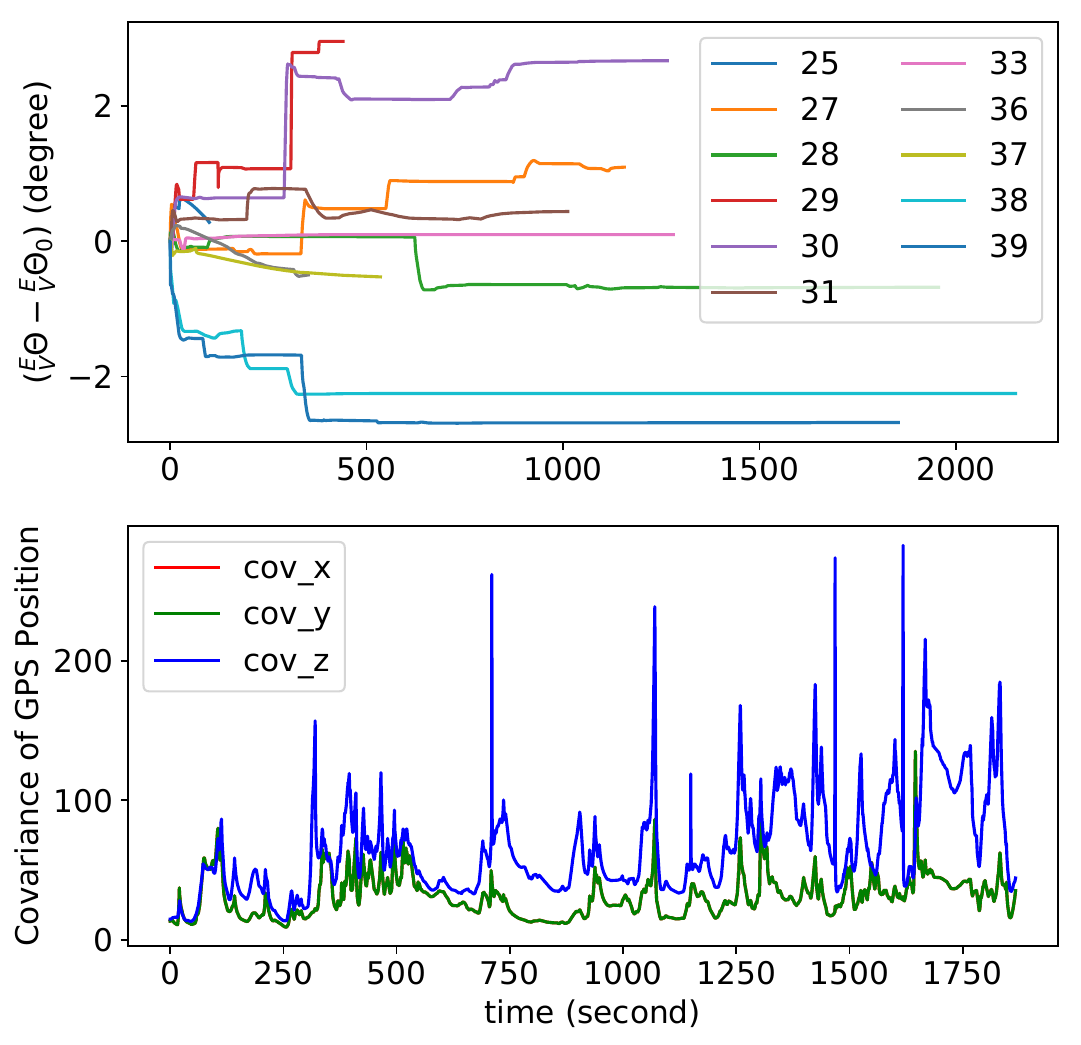}
  \caption{Top: $\left( {{}_V^E\theta  - {}_V^E{\theta _0}} \right)$  convergence over time. Bottom: An example of real world GPS position covariance in three directions. The values in the $x$ and $y$ directions are close.}
  \label{Yaw convergence and gps_cov}
\end{figure}
\begin{figure}[htbp]
  \centering
  \includegraphics[width=0.48\textwidth]{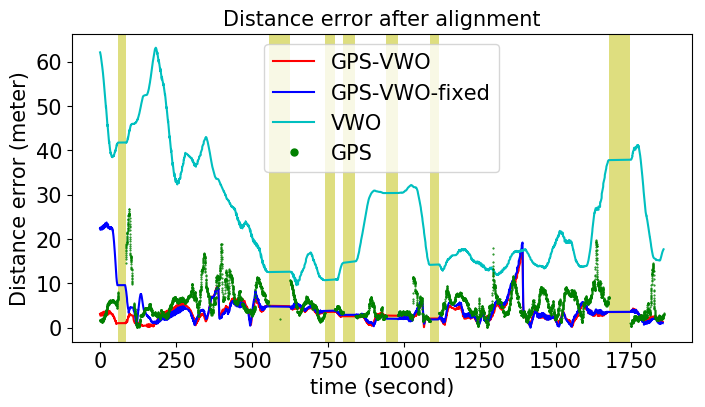}
  \caption{Distance error over time. The yellow shaded part means that the GPS is unavailable for at least 20s.}
  \label{Distance error of kaist}
\end{figure}

The initialization of ${}_V^ER$  and ${}^E{p_V}$ is solved by the SE(3) trajectory alignment of GPS and VWO. Pitch, roll and ${}^E{p_V}$ are fixed and only yaw is estimated online within the implementation of GPS-VWO. 
${}_V^E\theta $ is the corresponding yaw angle of ${}_V^ER$. In Fig.\ref{Yaw convergence and gps_cov}, we plot how $\left( {{}_V^E\theta  - {}_V^E{\theta _0}} \right)$ evolves on different datasets, where ${}_V^E{\theta _0}$ is the initial value of ${}_V^E\theta $ . ${}_V^E\theta $ on all datasets show the convergence trend over time, and the deviation from the initial value is less than ${3^ \circ }$. The Non-Gaussian behavior of real GPS position measurements is one of the reasons for the unsmooth convergence curves. As an example, the covariance of the GPS position provided by the KAIST Urban39 dataset is shown in the Fig.\ref{Yaw convergence and gps_cov}. 
Another reason is that the GPS signal is intermittent, as can be seen in the Fig.\ref{Distance error of kaist}.

We evaluate the ATE of GPS, VWO, GPS-VWO-fixed, VIO (Open-VINS \cite{geneva2020openvins}), GPS-VIO-fixed and GPS-VWO on all selected datasets. GPS-VIO-fixed is our implementation of \cite{lee2020intermittent} since it is not open-sourced. Results are summarized in Tab.\ref{table1}. 
Half of the sequences cannot be initialized by VIO. This is an expected phenomenon because the data collection vehicle was driven at approximately constant velocity. VIO initialization process suffers from this specific motion profile \cite{wu2017vins}\cite{liu2019visual}. The initialization of VWO is much more robust since velocity is directly obtained from the wheels' encoders.

We compare the accuracy of our VWO with  Visual-Inertial-Wheel Odometry (VIWO) \cite{lee2020visual}. \cite{lee2020visual} only presented the result on one dataset, Urban39. Therefore, this method is not displayed in Tab.\ref{table1}. Their best ATE was 42.748m. Thus we achieve better performance even IMU is not used.

\changemarker{
Meanwhile, we notice that the drifts of VWO on Urban36 and Urban37 are relatively larger than others in Tab.\ref{table1}. This is a known issue for wheel-odometer and also reported in the TABLE II of \cite{liu2021bidirectional}. Although the drifts can be significantly reduced by adjusting the odometer noise parameters, we retain these results for parameter consistency. They can also be seen as challenge cases when fusing with GPS.
}

It is clear from Tab.\ref{table1} that integrating GPS information can effectively restrict the drift of VWO or VIO, \changemarker{especially for challenge cases (Urban36 and Urban37).}
Meanwhile, GPS-VWO achieves the best localization accuracy because of the online estimation of the rotational extrinsic parameter, compared with GPS-VWO-fixed and GPS-VIO-fixed. 

\begin{figure}[htbp]
  \centering
  \includegraphics[width=0.48\textwidth, height=0.3\textwidth]{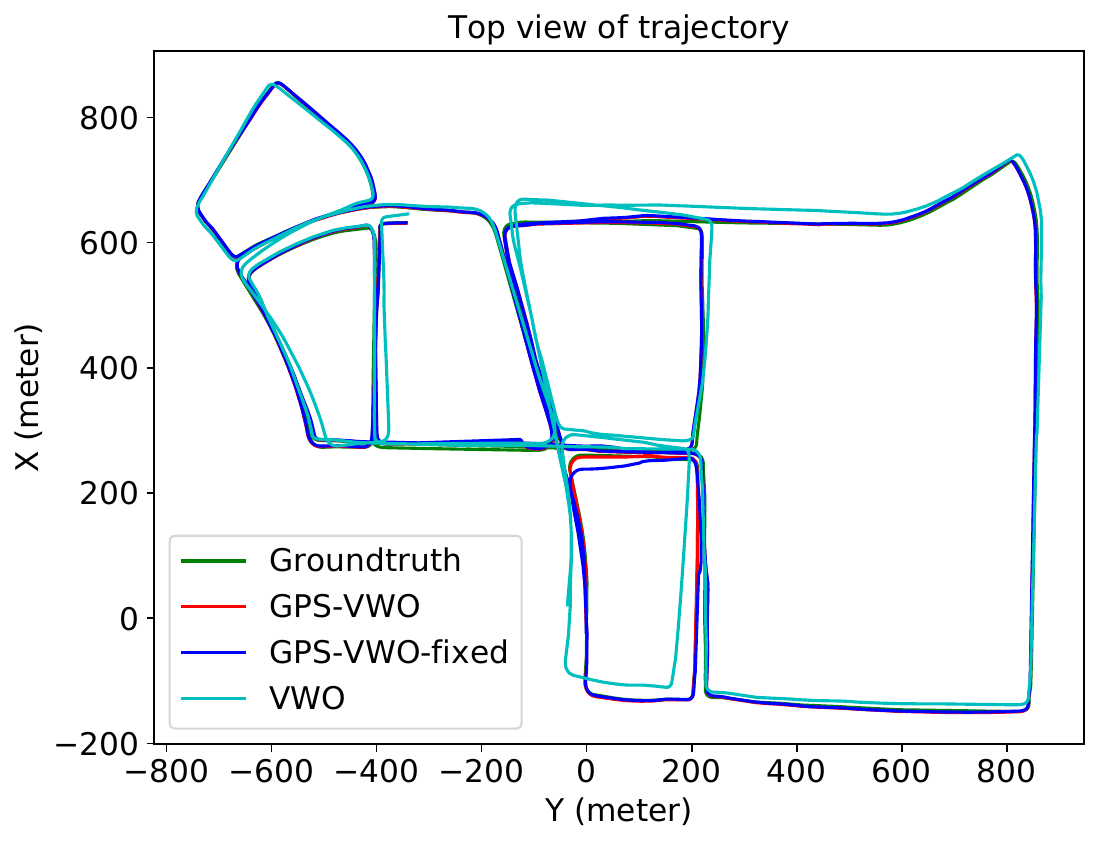}
  \caption{Comparison of aligned trajectories with different methods.}
  \label{traj_kaist}
\end{figure}

For a more intuitive presentation, aligned trajectories of the different localization methods on Urban39 dataset are shown in Fig.\ref{traj_kaist}. The trajectories of VIO and GPS-VIO-fixed are omitted as they are not the focus of this work, and removing them makes this figure easier to read. In Fig. \ref{Distance error of kaist}, the distance error after alignment is also presented. The yellow shaded part indicates that there is a GPS signal outage for at least 20s during this period. The longest GPS outage interval exceeds 70s. This shows the robustness of GPS-VWO under intermittent GPS measurements.

\begin{table}[ht]
\caption{ATE(meter) Comparison of Different Methods
on the KAIST Dataset}
\label{table1}
\begin{center}
\begin{tabular}{|c|c|c|c|c|c|c|c|}
\hline
\textbf{\makecell{ID}} & \textbf{\makecell{Path\\ len(km)}} & \textbf{\makecell{GPS}} 
& \textbf{\makecell{A}} & \textbf{\makecell{B}} & \textbf{\makecell{C}} & \textbf{\makecell{D}} & \textbf{\makecell{Ours}} \\
\hline
25 & 2.5 &  7.09 &  11.35 &	4.93 & $\times$ & $\times$ & \textbf{2.09} \\ \hline
27 & 5.4 &	9.97 &	22.61 &	6.84 & $\times$ & $\times$ & \textbf{6.79} \\ \hline
28 & 11.5 &	8.66 &	61.93 &	7.20 & 10.78 & 7.71 & \textbf{3.45} \\ \hline
29 & 3.6 &	9.63 &	16.89 &	8.74 & $\times$ & $\times$ & \textbf{7.92} \\ \hline
30 & 6.0 &	10.12 &	44.55 &	7.72 & $\times$ & $\times$ & \textbf{4.57} \\ \hline
31 & 11.4 &	7.26 &	49.32 &	4.58 & 76.87 & 6.85 & \textbf{3.78} \\ \hline
33 & 7.6 &	8.95 &	31.17 &	6.40 & $-$ & 7.77 & \textbf{2.67} \\ \hline
36 & 9.0 &	20.07 &	326.22 & 8.32 & $\times$ & $\times$ & \textbf{5.88} \\ \hline
37 & 11.8 &	6.18 &	156.64 & 5.13 & $\times$ & $\times$& \textbf{3.20} \\ \hline
38 & 11.4 &	7.09 &	55.36 &	4.94 & 7.53 & 5.53 & \textbf{3.75} \\ \hline
39 & 11.1 &	6.43 &	28.88 &	5.93 & 8.73 & 5.50 & \textbf{4.06} \\ \hline
\end{tabular}
\end{center}
\begin{tablenotes}
   \item $^{1}$ A, B, C, D and Ours represent VWO, GPS-VWO-fixed, VIO \cite{geneva2020openvins}, GPS-VIO-fixed \cite{lee2020intermittent} and GPS-VWO respectively.
   \item $^{2}$ $\times$ means VIO initialization is failed. $-$ means trajectory divergence.
\end{tablenotes}
\end{table}

\section{Conclusion}

We propose a novel MSCKF-based GPS-VWO system, which fuses camera, wheel encoder and GPS measurements in a tightly-coupled manner. The observability of the extrinsic parameter between the GPS global coordinate frame and the VWO reference frame is discussed to decide whether to include this parameter in the state vector for online refinement. The observability analysis concludes that the translation part of the extrinsic parameter is unobservable, while the rotation part is observable if it is a 1DoF parameterization. This observation proof is supported by the simulation results. 
\changemarker{
Real-world experimental results on multiple large-scale urban driving datasets demonstrate that the fusion of GPS and VWO provides better accuracy than GPS, 
}
and the online calibration of rotational extrinsic parameter further improves the localization accuracy of the estimator.

For future work, we would like to investigate the  combination of IMU and implement a GPS-VIWO system. 





\section*{APPENDIX}


\changemarker{The variance of unobservable state converges to zero is a non-trivial phenomenon, therefore we specifically explain here that this phenomenon is theoretically possible to occur.}

Here we present a simplified example to explain why the variance of ${}_V^E\theta $ could approach to 0 when ${}_V^E\theta $ is a 3DoF parameterization. Considering the following system model and measurement model:

\begin{equation}
    {}_V^E\dot \theta  = 0,{\rm{  }}{}^E{p_G} = {}_V^ER{}^V{p_G}
\end{equation}

Assuming ${}^E{p_G}$ and ${}^V{p_G}$ are direct measurements with Gaussian noise and the trajectory is a straight line:

\begin{equation}
    {}^E{p_G} = {\left[ {\begin{array}{*{20}{c}}
    {v_x}&{v_y}&0
    \end{array}} \right]^T}t
\end{equation}

Where $v_x$ and $v_y$ represent horizontal velocity. Thus, $H_\theta$ becomes (see equation (\ref{eq: 3dof_jacob})):

\begin{equation}
    \scalebox{0.9}{$
    {H_\theta } = \left[ {\begin{array}{*{20}{c}}
    0&0&{v_y}\\
    0&0&{-v_x}\\
    {-v_y}&{v_x}&0
    \end{array}} \right]t
    $}
\end{equation}

The covariance of ${}_V^E\theta $, denoted as $P$, should satisfy Riccati Equation:
\begin{equation}
    \scalebox{0.9}{$
    \begin{array}{l}
    \dot P =  - PH_\theta ^T{H_\theta }P\\
    H_\theta ^T{H_\theta } = \left[ {\begin{array}{*{20}{c}}
    {v_y^2}&{-v_x v_y}&0\\
    {-v_x v_y}&{v_x^2}&0\\
    0&0&{v_x^2 + v_y^2}
    \end{array}} \right]{t^2}
    \end{array}
    $}
\end{equation}

Measurement noise covariance $R$ is ignored here as it does not affect the analysis. The diagonal elements of $H_\theta ^T{H_\theta }$ are positive, therefore the variance of ${}_V^E\theta $ could approach to 0. 


\changemarker{
To validate the above derivation, we conduct simulation experiments as section \ref{Validation of the Observability Analysis}. $v_x$ and $v_y$ are set as 3 m/s and 4 m/s respectively.

Fig.\ref{appendix results} shows the calibration results of rotation extrinsic parameter ${}_V^E\theta $ when 1DoF parameterization and 3DoF parameterization are adopted for ${}_V^E\theta $. The state and corresponding variance convergence results are similar as in Fig.\ref{Yaw merge} and Fig.\ref{ypr merge}. ${}_V^E\theta $ is observable when it is parameterized with 1DoF, yet unobservable when it is parameterized with 3DoF. Results again support our observability analysis. Moreover, a novel theoritical finding is also demonstrated that the variance of unobservable state could converge to zero for some specific Kalman filter systems (see Fig.\ref{ypr merge} and Fig.\ref{appendix_ypr}).

\begin{figure}[htbp]
  \centering
    \begin{subfigure}[t]{0.235\textwidth}
        \centering
        \includegraphics[width=\textwidth, height=1.2\textwidth]{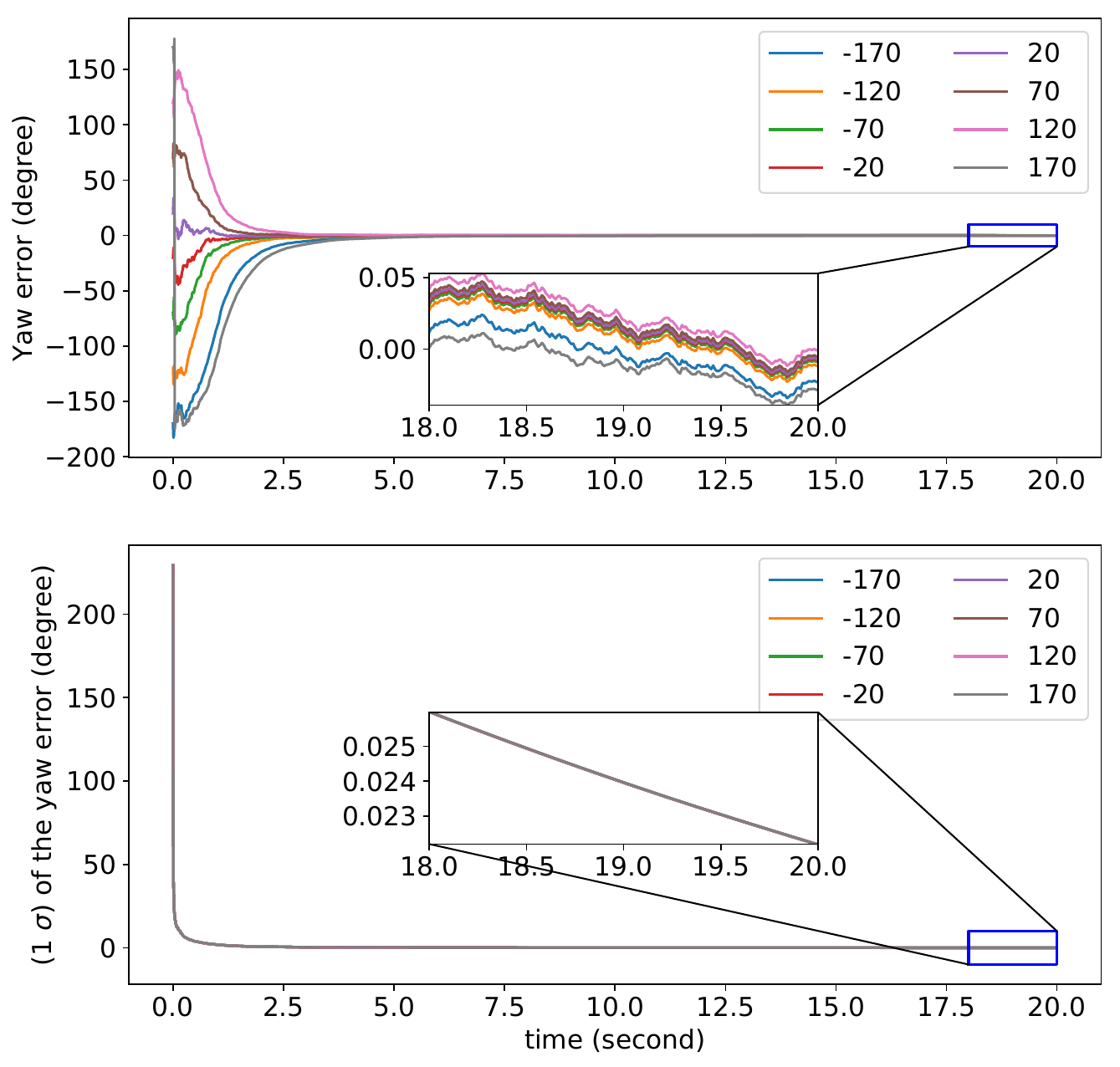}
        \caption{Top: Yaw convergence over time with respect to different initial guesses. Bottom: One standard deviation (1 $\sigma $) of yaw.}
    \end{subfigure}
    \hfill
    \begin{subfigure}[t]{0.235\textwidth}
        \centering
        \includegraphics[width=\textwidth, height=1.2\textwidth]{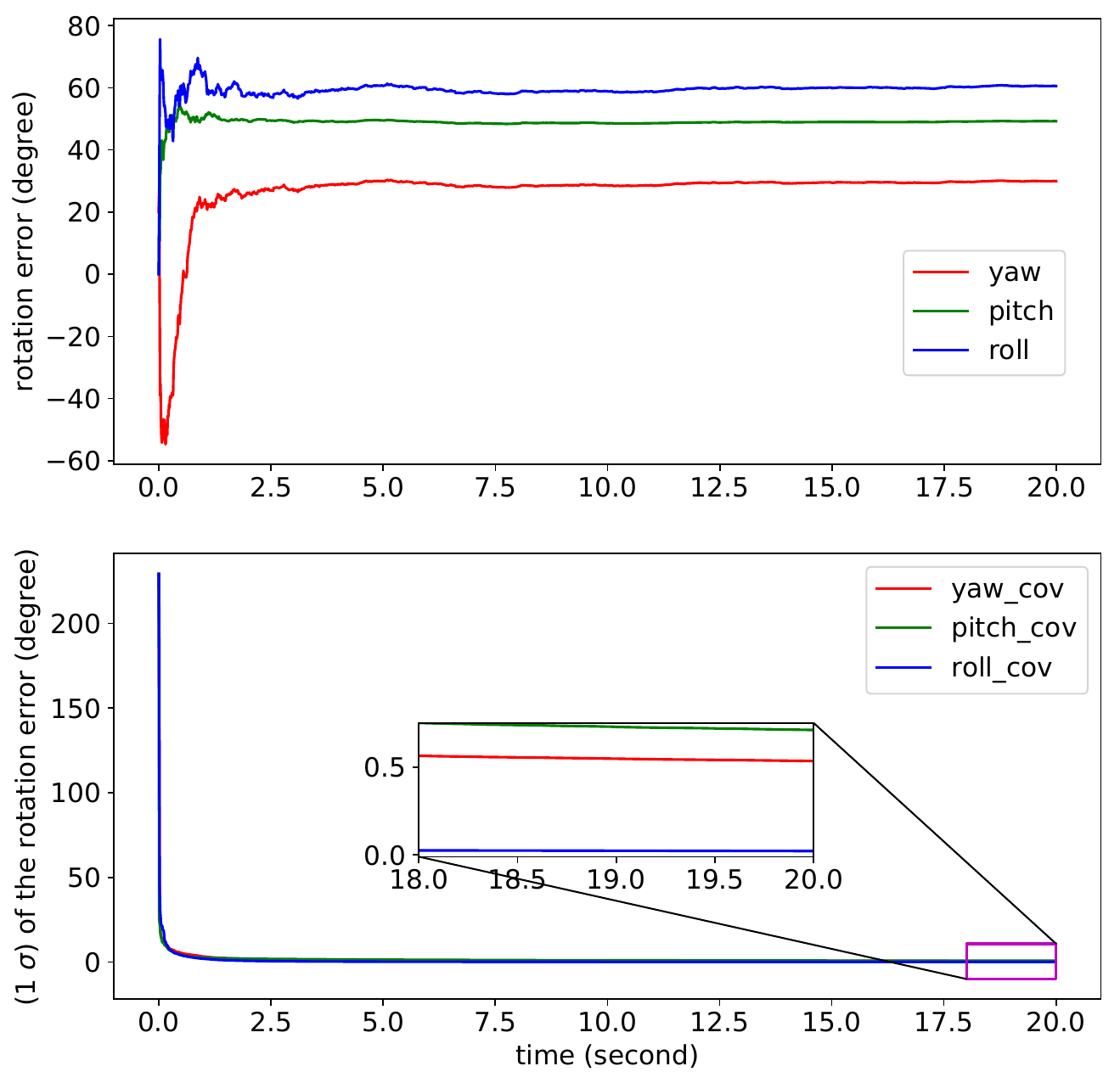}
        \caption{Top: rotation error over time. Bottom: One standard deviation (1 $\sigma $) of rotation error.}
        \label{appendix_ypr}
    \end{subfigure}
  \caption{Calibration results of different parameterization for rotation extrinsic parameter ${}_V^E\theta $.}
  \label{appendix results}
\end{figure}
}




\bibliographystyle{ieeetr}
\bibliography{bib}

\end{document}